\documentclass[acmsmall]{acmart}

\usepackage{algorithmic}
\usepackage[ruled,vlined]{algorithm2e}

\AtBeginDocument{%
  \providecommand\BibTeX{{%
    \normalfont B\kern-0.5em{\scshape i\kern-0.25em b}\kern-0.8em\TeX}}}

\setcopyright{acmcopyright}
\copyrightyear{2018}
\acmYear{2018}
\acmDOI{10.1145/1122445.1122456}

\acmJournal{JACM}
\acmVolume{37}
\acmNumber{4}
\acmArticle{111}
\acmMonth{8}

\begin{document}

\title{LW-GCN: A Lightweight FPGA-based Graph Convolutional Network Accelerator}

\author{Zhuofu Tao}
\authornote{Both authors contributed equally to this research.}
\email{z24tao@g.ucla.edu}
\author{Chen Wu}
\authornotemark[1]
\email{chenwu1989@g.ucla.edu}

\author{Yuan Liang}
\email{liangyuandg@g.ucla.edu}

\author{Lei He}
\email{lhe@ee.ucla.edu}

\affiliation{%
  \institution{Electrical and Computer Engineering, University of California, Los Angeles}
  \streetaddress{Westwood}
  \city{Los Angeles}
  \state{CA}
  \country{USA}
  \postcode{90095}
}

\renewcommand{\shortauthors}{Zhuofu Tao and Chen Wu, et al.}

\begin{abstract}
Graph convolutional networks (GCNs) have been introduced to effectively process non-euclidean graph data. However, GCNs incur large amounts of irregularity in computation and memory access, which prevents efficient use of traditional neural network accelerators. Moreover, existing dedicated GCN accelerators demand high memory volumes and are difficult to implement onto resource limited edge devices.

In this work, we propose LW-GCN, a lightweight FPGA-based accelerator with a software-hardware co-designed process to tackle irregularity in computation and memory access in GCN inference. LW-GCN decomposes the main GCN operations into sparse-dense matrix multiplication (SDMM) and dense matrix multiplication (DMM). We propose a novel compression format to balance workload across PEs and prevent data hazards. Moreover, we apply data quantization and workload tiling, and map both SDMM and DMM of GCN inference onto a uniform architecture on resource limited hardware. Evaluation on GCN and GraphSAGE are performed on Xilinx Kintex-7 FPGA with three popular datasets. Compared to existing CPU, GPU, and state-of-the-art FPGA-based accelerator, LW-GCN reduces latency by up to 60x, 12x and 1.7x and increases power efficiency by up to 912x., 511x and 3.87x, respectively. Furthermore, compared with NVIDIA’s latest edge GPU Jetson Xavier NX, LW-GCN achieves speedup and energy savings of 32x and 84x, respectively. 
\end{abstract}



\maketitle

\section{Introduction}
Over recent years, deep learning paradigms such as convolutional neural networks (CNNs) and recurrent neural networks (RNNs) have shown great success in various families of tasks such as image and text processing \cite{cite-image-processing, cite-text-processing}. However, these paradigms rely heavily on structural properties of euclidean data such as dense tensors, and have trouble processing non-euclidean data such as graphs. To tackle this problem, graph neural networks (GNNs) have been introduced and have demonstrated the ability to accurately process complex graph data~\cite{cite-gnn}. Among numerous GNNs, graph convolutional networks (GCNs) \cite{cite-gcn}, which borrows ideas from CNNs to aggregate neighbor data, have quickly attracted industrial attention as a popular solution to real-world problems~\cite{cite-drug-development, cite-action-recognition, cite-recommender, cite-graph-clustering}. Since then, there have been many other graph processing algorithms ({\it i.e.} GIN, GraphSAGE, GAT, etc.) introduced to optimize performance on existing problems and extend to new challenges \cite{cite-gin, cite-graphsage, cite-gat, cite-agnn, cite-rgcn, cite-splinecnn}.

Similar to CNN, GCNs also contain multiple layers, where the main operations of each layer are \textit{combination} and \textit{aggregation}. \textit{Combination} is similar to a dense layer of a multi-layer perceptron (MLP), where a feature matrix is multiplied by a weight matrix.  \textit{Aggregation} is similar to a convolution operation of a standard CNN, where the feature vector of each vertex is computed through a weighted aggregation of all feature vectors of neighboring vertices, which can be represented as a matrix multiplication between the graph adjacency matrix and the feature matrix.

Despite the fact that the majority of GCN operations can be represented as matrix multiplication, it is unlikely for existing matrix multiplication oriented accelerators~\cite{cite-opu, cite-tpu, cite-light-opu, cite-cambricon} to yield high throughput on GCN. These accelerators typically exploit the structured nature of dense tensors and apply data reuse techniques to achieve performance boosts. However, such techniques are ineffective in GCNs because adjacency matrices in GCNs are often sparse, random, and irregular due to the fact that node degree distribution of random graphs follow the power law distribution. Although existing works such as EIE and Cambricon-X~\cite{cite-eie, cite-cambricon-x} tackles irregularity in computation and memory access in deep compressed CNNs, the sparsity of deep compressed CNNs is much lower (around 90\%) than that of GCNs (over 99.9\%). Due to the extreme sparseness of graph data, sparse CNN accelerators also fail to maximize computational efficiency, thus a more effective approach is required.

There are existing GCN accelerators to overcome the sparseness challenges~\cite{cite-hygcn, cite-awb-gcn, cite-engn}. Although they achieve performance boosts, they either cache large amounts of data on-chip or rapidly load data from off-chip memory. This requires either large amounts of on-chip memory or huge off-chip memory bandwidth ({\it i.e.} high-bandwidth memory (HBM)). In addition, previous work tends to employ hardware designs with unnecessary components on top of reliance on powerful hardware capability. For example, HyGCN~\cite{cite-hygcn} and AWB-GCN~\cite{cite-awb-gcn} assume {\it combination} and {\it aggregation} are structurally different and deploy independent hardware modules for each operation. Despite the efforts to balance computation in each module, the inherent workload differences across datasets make it difficult to keep both modules fully utilized. Moreover, as GCN grows in popularity and supports numerous real-world applications, it is natural for its inference workload to see heavy demand on edge devices in the near future. It is unlikely for these resource limited devices to provide powerful hardware resources, therefore a more lightweight approach is required.

To this end, we propose a lightweight software-hardware co-optimized accelerator, named {\bf LW-GCN}, to efficiently perform GCN inference. We first introduce the "packet" conception in compressing the sparse matrix into a packet-level column-only coordinate-list (PCOO) format in software. The PCOO format is also easy to decompress in the hardware. We then propose a unified micro-architecture to efficiently execute both {\it combination} and {\it aggregation}, where the main operations are dense matrix multiplication (DMM) and sparse-dense matrix multiplication (SDMM). An optimized computation pipeline is utilized in each processing element (PE) to cope with the irregularity in computation and memory access caused by SDMM. Due to the limited hardware resources, we apply tiling to process a portion of DMM/SDMM at a time, which enables us to only keep a fraction of the matrices on-chip. Finally, our preprocess procedure injects "empty elements" in PCOO to indicate idle cycles and prevent data collisions caused by irregularity of the sparse matrix in software side. The preprocess algorithm has linear time and space complexity with respect to the number of elements in the sparse matrix.

\begin{table*}[tb]
    \centering
    \footnotesize
    \caption{Dimensions and densities of widely-used datasets.}
    \label{dataset-details}
    \begin{tabular}{l|c|c|c|c|c|c|c} \hline\hline
        Datasets & Nodes & Edges & Input Features & Classes & Feature Density & Edge Density & Weight Density \\ \hline
        Cora     &2708  &10556 &1433 &7 &1.27\% &0.144\%  &100\% \\ \hline
        CiteSeer &3327  &9104  &3703 &6 &0.85\% &0.0822\% &100\% \\ \hline
        PubMed   &19717 &88648 &500  &3 &10.0\% &0.0228\% &100\% \\ \hline \hline
    \end{tabular}
\end{table*}

We implement \textbf{LW-GCN} onto the Xilinx Kintex-7 K325T FPGA, which simulates the limited resource availability of edge devices. We evaluate \textbf{LW-GCN} for GCN and GraphSAGE on three popular datasets Cora~\cite{cite-cora}, CiteSeer~\cite{cite-citeseer} and PubMed~\cite{cite-pubmed}. Compared to state-of-the-art software framework Pytorch Geometric (PyG) running on Intel Xeon Gold 5218 CPU, NVIDIA Jetson Xavier NX edge GPU, NVIDIA RTX3090 GPU, and a prior FPGA-based GCN accelerator~\cite{cite-awb-gcn}, {\bf LW-GCN} achieves up to 60$\times$, 32$\times$, 12$\times$, and 1.7$\times$ smaller latency, as well as 912$\times$, 84$\times$, 511$\times$, and 3.87$\times$ higher energy efficiency, respectively. To summarize, the main contributions of this work as listed as:

\begin{itemize}
    \item {\bf Software-Hardware Co-optimization.} We propose a linear time and space preprocess algorithm to compress the sparse matrix into PCOO format and optimize the GCN workload. In addition, the micro-architecture is designed to efficiently process the PCOO format, so that the GCN workload is also optimized in hardware side.  
    
    \item {\bf High Computation Efficiency.} We design unified micro-architecture for DMM and SDMM, which efficiently performs both {\it combination} and {\it aggregation} operations in GCN. Moreover, the PCOO format skips computation and storage of zeros in the sparse matrix, and the optimized architecture in each PE addresses the irregularity issue caused by sparse matrix, which further increase the computation efficiency of {\bf LW-GCN}. 
    
    \item {\bf Low Resource Requirement.} The compression method in the preprocess algorithm reduces both the storage and bandwidth requirement. Moreover, {\bf LW-GCN} utilizes tiling techniques to process a portion of DMM/SDMM at a time, thus further alleviating on-chip memory burdens. Different from prior works that rely heavily on large on-chip memory availability, {\bf LW-GCN} works effectively on resource limited edge devices.
    
    \item {\bf High Performance.} We evaluate {\bf LW-GCN} on a Kintex-7 FPGA on three popular datasets. Our work reduces latency by up to 60$\times$, 32$\times$, 12$\times$, and 1.7$\times$ and increases energy efficiency by up to 912$\times$, 84$\times$, 511$\times$, and 3.87$\times$, compared to Intel CPU, NVIDIA edge GPU, NVIDIA server GPU and prior FPGA-based GCN accelerator.
\end{itemize}

\section{Challenges and Motivations}
In this section, we will briefly introduce the GCN algorithm, the challenges to map it on hardware, and the motivation of our accelerator design.

\subsection{GCN Background}
\label{gcn_background}
The forward propagation of the $l$th layer of a multi-layer GCN~\cite{cite-gcn} is illustrated in equation~(\ref{gcn-layer}),

\begin{align}
\label{gcn-layer}
    X_l = Relu(AX_{l-1}W_l),
\end{align}

where $A, X_{l}$ and $W_l$ indicate the adjacency matrix of the input graph, the feature matrix of the $l$th layer, and the weight matrix of the $l$th layer, respectively. $Relu$ is the activation function and the input feature matrix of the graph is represented as $X_0$. 

Based on our analysis on widely-used datasets, the adjacency matrices are often sparse, the input feature matrix is often sparse, while the weight matrices are dense, as shown in Table~\ref{dataset-details}. Therefore, the computation order influences dramatically on computation complexity when skipping the zeros. Following the analysis in~\cite{cite-awb-gcn}, we profile the required number of scalar operations and intermediate storage under different computation orders, as shown in Table~\ref{gcn-complexity}. This way, we perform $A \times (X_{l-1} \times W_l)$ as it is much more efficient. For simplicity, we refer to step $X_{l-1} \times W_l$ as \textit{combination} and $A \times (...)$ as \textit{aggregation} of each GCN layer, following the conventions of \cite{cite-hygcn}. Moreover, for the {\it combination} of the first layer and {\it aggregation}, we perform SDMM and for the {\it combination} of other layers, we perform DMM, this is because $X_l$ is produced by the previous layer and it is always dense except the first layer. From here on, for SDMM we will refer to the left sparse input matrix as $X$, the right dense input matrix as $W$, and output matrix as $Y$ for simplicity.

\begin{table}[tb]
    \centering
    \caption{Required computation and storage under different computation orders.}
    \label{gcn-complexity}
    \begin{tabular}{l|c|c} \hline \hline
        Datasets & $(A \times X_{l-1}) \times W_l$ & $A \times (X_{l-1} \times W_l)$ \\
        \hline
        Cora     &18.7M / 56.2Mb &1.33M / 0.661Mb \\ \hline
        CiteSeer &38.9M / 188Mb &2.23M / 0.812Mb \\ \hline
        PubMed   &118M / 150Mb &18.6M / 4.81Mb \\ \hline \hline
    \end{tabular}
\end{table}

\subsection{Challenges}
\label{challenges}

As illustrated in previous sections, the main operations in GCN can be extracted as SDMM and DMM. Therefore, the challenge is to accelerate SDMM and DMM on resource limited devices.

\subsubsection{Challenges on SDMM}
The computation of SDMM on one PE can be effective to skip all zero elements of the sparse input $X$, as shown in Algorithm~\ref{SDMM-algorithm}. However, parallel computing with multiple PEs introduce new problems in {\bf Computation Imbalance} and {\bf Memory Irregularity}. 

\begin{algorithm}[tb]
    \SetAlgoLined
    \textbf{inputs:} $X \in \mathbb{R}^{m \times n}$, $W \in \mathbb{R}^{n \times p}$, $Y \gets 0^{m \times p}$\;
    \For{$X_{i,j}$ in $X$}{
        \If{$X_{i,j} \ne$ 0}{
            \For{$W_{j,k}$ in $W$} {
                $Y_{i,k} \gets Y_{i,k} + X_{i,j} \times W_{j,k}$\;
            }
        }
    }
    \textbf{return} $Y$
    \caption{SDMM}
    \label{SDMM-algorithm}
\end{algorithm}

{\bf Computation Imbalance:} To accelerate the computation of SDMM on multiple PEs, we will first divide the workload and distribute portions to multiple PEs. In each PE, we only process the non-zero elements from $X$. Due to irregularity in $X$, it is difficult to allocate identical workloads to every PE, which leads to computation imbalance. This is challenging for the SDMM in GCN, as the matrices in {\it combination} and {\it aggregation} are extremely sparse ($>99\%$). Moreover, real-world graphs follows the power law distribution~\cite{cite-power-law-gcn}, which implies that the minority of rows (columns) in the adjacent matrix have the majority of non-zeros while the majority of rows have only a few (not empty) non-zeros. Such irregularity further increases the difficulty to balance workload.

{\bf Memory Irregularity:} Since the optimization of SDMM only stores the non-zero elements to save memory requirement, the data irregularity incurs several issues during computation. Firstly, it is difficult to predict the position of the next non-zero element $X_{i, j}$ to be processed in the left matrix. Since matrix multiplication matches $X_{i,j}$ against $W_j$ and $j$ is unknown, the next non-zero $X_{i,j}$ could require any row of $W$. This uncertainty requires us to cache the entire $W$ matrix on-chip, which leads to very expensive caching. Secondly, parallel computing of SDMM will process multiple non-zero elements of $X$ simultaneously, thus requiring all corresponding data in $W$ to be readily available, this introduces the problem of bank conflict. For example, to process non-zero elements $X_{i_a,j_a}$ and $X_{i_b,j_b}$ simultaneously, the PEs must be provided with $W_{j_a}$ and $W_{j_b}$. However, memory resources on FPGA usually come with high depth and very limited (1 or 2) ports, where each port can only access a single depth of the memory bank at a time. In the scenario where $W_{j_a}$ and $W_{j_b}$ are stored on the same bank, which can only supply one of them at a time, we face a data conflict. Thirdly, since the SDMM algorithm computes each row of the SDMM result as a sum of many scalar-vector multiplications, it introduces a read-after-write (RAW) conflict. This is due to the fact that arithmetic operations tend to take multiple cycles on hardware. If we process non-zero elements $X_{i, j_a}$ followed by $X_{i, j_b}$ in the immediate next cycle, the multiplication and addition would not have finished in the first cycle. When the PE reads in $Y_i$ in the next cycle to process addition for $X_{i, j_b}$ it would inevitably read in an incorrect result. Finally, although the RAW conflict can be effectively resolved by utilizing multiply-accumulators (MACs) instead of individual multipliers and adders, doing so restraints the design to use the same PE to process each row $X_i$, which leads back to the issue of {\bf Computation Imbalance}. As the individual node degree in a random graph follows the power law distribution, it is common for there to be a large difference (over 100$\times$) between densities of individual rows of an adjacency matrix. Naively partitioning the sparse input $X$ into row-blocks and assigning row-blocks to a PE group would result in a difference between non-zero workload assigned to each PE within the group. The latency of the group would be controlled solely by the input row with the highest density, vastly reducing efficiency.

\subsubsection{Challenges on resource limited devices}
Accelerating the inference of GCN should include the acceleration of both DMM and SDMM. Although DMM does not have the issues of {\bf Computation Imbalance} and {\bf Memory Irregularity} as SDMM, DMM requires storage of all the numbers in the matrices, which incurs the issue of {\bf Bandwidth Constraints}. Therefore, designing a module with both DMM and SDMM in consideration is challenging. Existing solutions such as \cite{cite-hygcn, cite-awb-gcn} view DMM and SDMM as inherently different workloads, therefore introduced dedicated modules to perform each independently. Although this allows each module to efficiently tailor toward its workload, the resource allocation for each module raises a non-negligible concern. Since different problem settings come with different data dimensions and densities (examples shown in Table \ref{dataset-details}), the ratio between arithmetic operations required in \textit{combination} and \textit{aggregation} varies significantly across datasets. In order to fully utilize the dedicated modules for each, these accelerators often need to dynamically allocate computation resources to each module for each problem setting, which consumes many hours or even days for the synthesis and implementation process. Moreover, the data dependency between {\it combination} and {\it aggregation} leaves one of the DMM and SDMM modules idle, which leads to a waste of resources. Such problem makes the accelerating GCN on resource limited devices more challenging.

\subsection{Motivation}
\label{motivation}
Motivated by the above challenges, we propose a software-hardware co-optimization process to address each of them, while keeping an available resource budget of an edge device. We first define a PCOO format to compress the input sparse matrix, effectively eliminating zero elements to preserve both storage space and computation time. We then design a dedicated computation engine processing multiple non-zero elements in parallel efficiently. Some key highlights of our design include:
\begin{itemize}
    \item {\bf Software Preprocessing:} We first compress the sparse data into PCOO format, and leverage the binary "edge-or-no-edge" feature of graph adjacency matrices to remove value data. Then, we search the space of the sparse matrix to balance the workload on different PEs in order to resolve the issue of computation imbalance. Finally, idle data insertion is applied to solve the problem of bank conflict with a small burden.
    \item {\bf Dedicated Architecture Design:} We design a dedicated architecture to decompress the PCOO format in order to further increase the computation efficiency. Moreover, a multi-port memory is applied in our design to resolve the issue of data conflict from the hardware side.
    \item {\bf Unified Micro-architecture:} We observe that DMM is essentially a special case of SDMM where density is 1. Therefore, we design the algorithm to process SDMM by individual non-zero elements on the sparse matrix, and apply that algorithm on DMM as well. Moreover, we design a unified architecture of PE to process both DMM and SDMM efficiently, which allows all computation resource to be fully utilized. This allows the full GCN workload to be deployed onto a unified module, resolving the resource allocation problem.
    \item \textbf{Flexible Design:} Our design is not dedicated toward any specific GCN configurations, instead it is able to support any number of layers with any size of GCN layers. Additionally, since DMM and SDMM are widely used across GNNs, our design supports most operations needed for many other networks. In Section~\ref{evaluation} we also evaluate our design on GraphSAGE in addition to GCN as we support it out of the box.
\end{itemize}

\section{Software Preprocessing}
\label{preprocessing}
The software preprocessing algorithm will first compress the input. data, then allocate and schedule GCN workloads onto different PEs. We will explain these algorithms in detail in this section. 

\begin{algorithm}[tb]
    \SetAlgoLined
    \textbf{inputs:} $X \in \mathbb{R}^{m \times n}$, $T$, $K$\;
    tiles, sor, eor, vld = [], $T \times 4$, $T \times 2$, $T$\;
    \For{$t \gets 0$ to $n-1$ by $T$}{
        rows $\gets$ [[] for 0:$K$]\;
        \For{$i \gets 0$:$m$}{
            row $\gets$ []\;
            \For{$j \gets t$:$(t + T - 1)$}{
                \If{$X_{i,j} \ne 0$}{
                    row.append($j$ \% $T$ + vld)\;
                }
            }
            row $\gets$ [0] if row is empty else row\;
            row[0] $\gets$ row[0] + sor\;
            row[-1] $\gets$ row[-1] + eor\;
            rows[$i$ \% $K$].extend(row)\;
        }
        fill zeros until rows is rectangular\;
        tiles.append(rows.transpose())\;
    }
    \textbf{return} tiles\;
    \caption{Sparse matrix preprocessing}
    \label{preprocessing-algorithm}
\end{algorithm}

\subsection{Data compression}

\subsubsection{PCOO format}

\begin{figure*}[tb]
    \centering
    \includegraphics[width=0.9\textwidth]{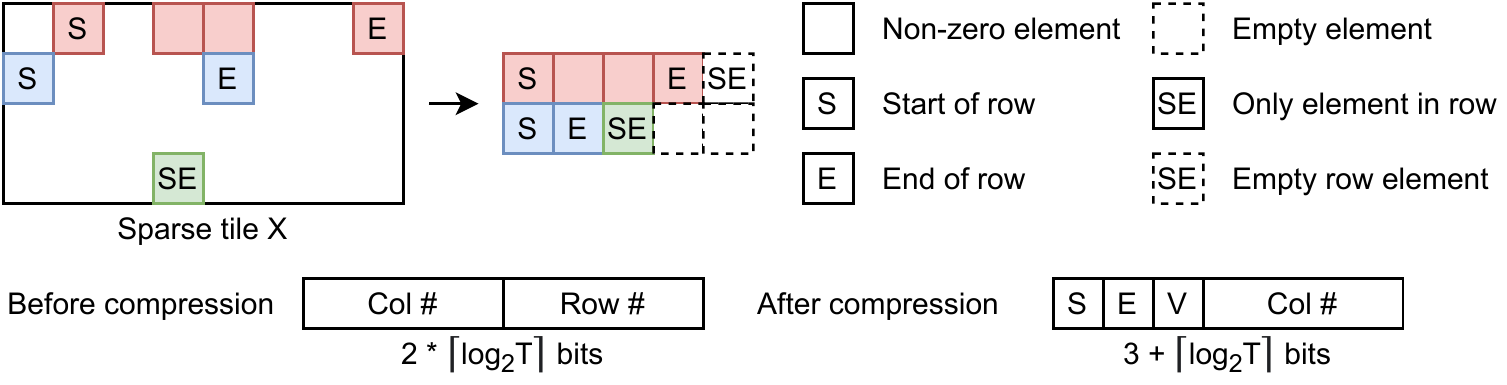}
    \caption{Packet-level column-only coordinate list format}
    \label{pcoo}
\end{figure*}

As shown in Table~\ref{dataset-details}, the adjacency matrix and the input matrix of the first layer in GCNs are often extremely sparse. Therefore, we compress these matrices to process only valuable information (non-zero elements) to save storage and reduce computation complexity. We introduce the "packet" concept to propose a packet-level column-only coordinate-list (PCOO) format to compress the sparse matrix (Fig. \ref{pcoo}). In detail, we treat all the elements in one row as one packet, and each non-zero element $X_{i,j}$ in one row is formatted into a bit-wise format. Firstly, the leading two bits conclude the row information of each non-zero element, which indicate the start-of-row (SOR) for first non-zero element and end-of-row (EOR) for last non-zero element. Secondly, the following one bit indicates valid (VLD) to differentiate from injected empty elements (the injected empty elements are explained in detail in Section~\ref{assignment_scheduling}). These three bits act as the header of a packet and the rest bits play a role of payload, which has the column information and the value of each non-zero element. We use $log_2(T)$ bits, where $T$ is the tile size, to represent the column position within tile ($j$ mod $T$) of $X_{i, j}$. Finally, we use the remaining $H$ bits to represent the value of the non-zero element. In the corner case where there are no non-zero elements in a given row, we set the header SOR = EOR = 1 and VLD = 0 with empty payload, in order to instruct the hardware to increment the row number without performing calculation. In this way, we totally need $3+log_2(T)+H$ bits to represent each non-zero element in the sparse matrix. The algorithm of compressing sparse matrix with PCOO is concluded in Algorithm~\ref{preprocessing-algorithm}.

Since we treat DMM the same operation as SDMM, we format the left dense matrix in DMM to fit the unified PE (as expressed in Section~\ref{sec:micro_arch}). The dense matrix is first stored as normal, and then we design all rows to share the same column information in PCOO format. In this way, we only need an extra of $(3+log_2(T)) \times Column\_Size$ bits to store the dense matrix in intermediate steps.

\subsubsection{Quantization}
In order to further reduce the memory requirement, we apply quantization onto the values of all the matrices in GCNs. After exploring the structure of GCN and GraphSAGE on three popular used datasets, we use 4-bit signed integer (SINT4) to quantize the values of the input features. In fact, for all matrices as well as two out of three feature matrices, the value of each $X_{i,j}$ would be binary between 0 and 1, and there would be no accuracy loss at all. During computation, we store the intermediate results as 32-bit signed integer (SINT32) and quantize them to SINT16 after obtaining the final results of each layer. Evaluated on both GCN and GraphSAGE on all three datasets, using our proposed quantization strategy incurs negligible accuracy loss of within 0.2\%. 



\begin{figure*}[tb]
    \centering
    \includegraphics[width=0.9\textwidth]{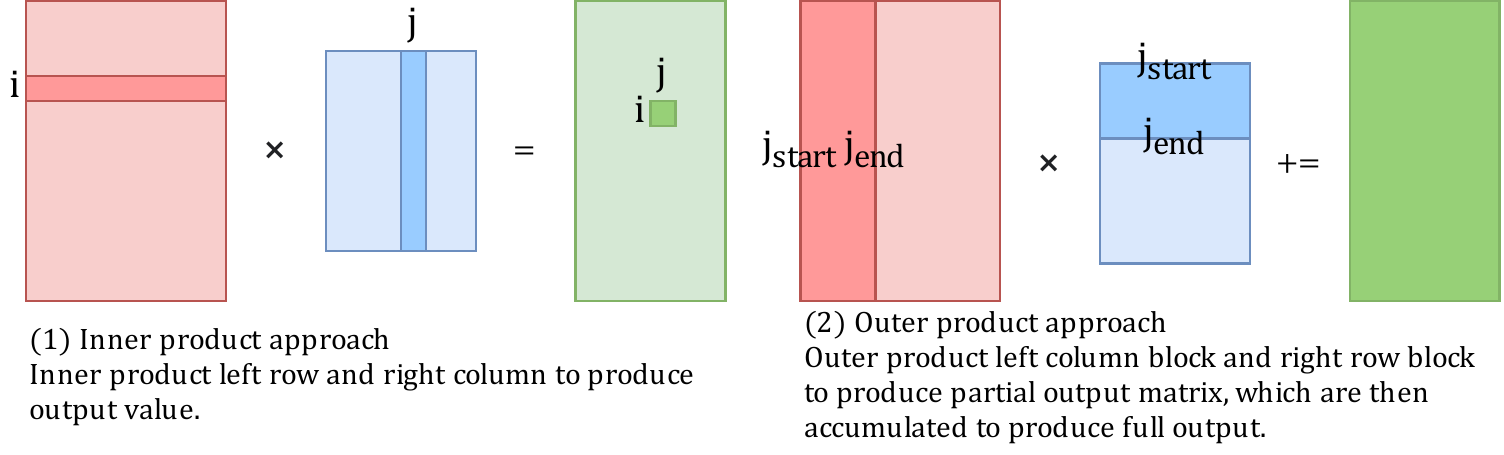}
    \caption{Outer product matrix multiplication}
    \label{tiling}
\end{figure*}

\subsection{Assignment and Scheduling}
\label{assignment_scheduling}

In order to reduce memory consumption, we employ an outer-product tiling approach, as shown in Fig.~\ref{tiling}. We partition the inputs into $T$-column tiles for $X$ and $T$-row tiles for $W$. The hardware processes a pair of tiles at a time, and produces the final result by accumulating all tile results. For each pair of tiles, we perform the following preprocessing steps to balance workload and reduce data volume.

\subsubsection{Workload assignment and scheduling} 
\begin{figure*}[tb]
    \centering
    \includegraphics[width=0.9\textwidth]{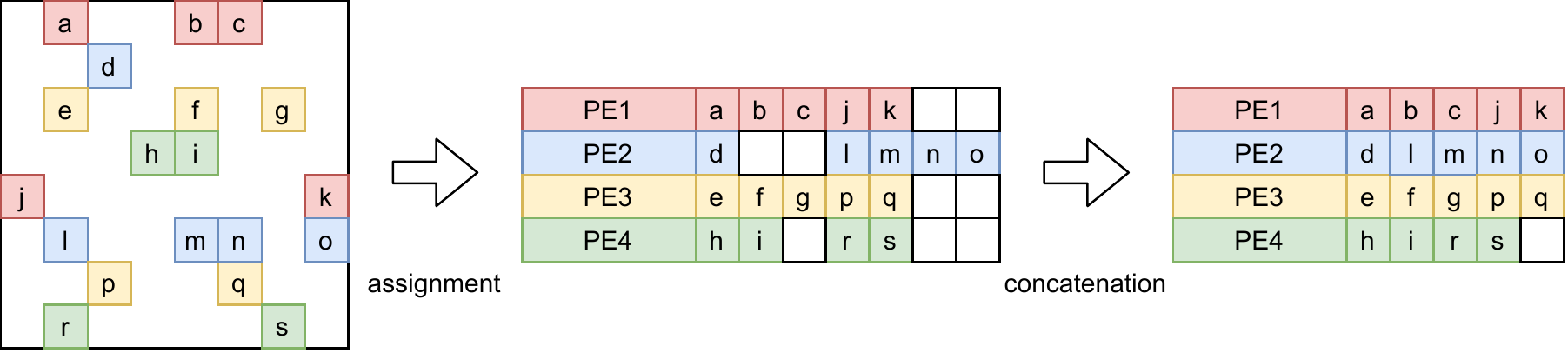}
    \caption{Round-robin assignment of non-zero elements to four PEs}
    \label{sparse-assignment}
\end{figure*}

Multiplication of non-zero elements in one row of the sparse matrix $X$ is assigned to the same PE, while multiplication of different rows are assigned to different PEs in a round-robin fashion. This way, non-zero elements from each row are processed sequentially on the same PE and do not require the same accumulator simultaneously. However, different rows of a graph adjacent matrix could have extremely different densities (with relative difference $> 100 \times$). If we naively tile the workload further into row blocks, it would be inefficient for the majority of PEs to finish execution and remain idle to wait for a single PE to finish processing a particularly dense row, shown as the assignment step in Fig.~\ref{sparse-assignment}. To increase PE efficiency, we design the PEs to work independently, each PE starts to compute a new row immediately when it finishes the previous one. This way, multiple rows are effectively concatenated before assigned to one PE, this way we eliminate idle time (shown as concatenation step in Fig.~\ref{sparse-assignment}). Since it is unlikely for the density of a row to correlate with its row number, by Law of Large Numbers we expect the sum of densities of rows assigned to each PE to be similar. In section \ref{evaluation}, we will analyze examples in details and compare the computation cost and idle time before and after the concatenation step. Finally, to ensure all PEs process the same amount of non-zero elements, we inject empty elements at the end of each concatenated row when necessary.

\subsubsection{Data collision resolution} Due to constraints of on-chip memory, multiple rows of $W$ are stored in the same memory slice, out of which only a single row may be accessed at any time. However, the sparsity of $X$ may cause two PEs to simultaneously access different depths on the same memory slice, which incurs data collision. To resolve this problem, we first develop a multi-bank memory system with data replication to reduce the occurrence of such data collisions (see details in Section~\ref{subsec:ddm_par}). We then inject an empty element with VLD = 0 to prevent any data collision not resolved by the multi-port memory system. The trade-off is made between the usage of on-chip memory and extra latency incurred by empty elements, detailed analysis will be discussed in Section\ref{latency-breakdown}.

Preprocessing is summarized in two steps in Algorithms \ref{preprocessing-algorithm} and \ref{collision-stalling}. Overall, this preprocessing algorithm is bounded by linear time and space complexity to the total number of non-zero elements in every unique sparse tile. On the other hand, the dense tile is quantized to SINT4 and passed to hardware without structural change. Finally, the preprocessor generates instructions to serialize the execution across layers and steps. 

\begin{algorithm}[tb]
    \SetAlgoLined
    \textbf{inputs:} $tile \in \mathbb{R}^{N \times K}$, depth $d$=16\;
    used, row $\gets$ [0 for 0:$K$], [-1 for 0:$K$]\;
    result, share, block, j $\gets$ [], \{\}, \{\}, 0\;
    \While{sum(used) $< N \times T$}{
        i $\gets$ used$_j$\;
        \uIf{$tile_{i,j}$ $\in$ share \textbf{or} $tile_{i,j}$ \% d $\not\in$ block}{
            row$_j \gets$ $tile_{i,j}$\;
            share.append($tile_{i,j}$)\;
            block.append($tile_{i,j}$ \% $d$)\;
            used$_j \gets$ used$_j$ + 1\;
        }
        \Else{
            row$_j \gets$ 0\;
        }
        \If{min(row) $\ne$ -1}{
            result.append(row)\;
            row, share, block $\gets$ [-1 for 0:$K$], \{\}, \{\}\;
        }
        j $\gets$ (j + 1) \% K\;
    }
    fill zeros until result is rectangular\;
    \textbf{return} result\;
    \caption{Collision stalling}
    \label{collision-stalling}
\end{algorithm}

\section{Micro-architecture of LW-GCN}
\label{sec:micro_arch}
As shown in Fig.~\ref{fig:over_arch}, the micro-architecture of {\bf LW-GCN} is composed of {\it Peripheral Interface, External Memory Interface, Top Control, PE Array for Sparse-Dense Matrix Multiplication} and on-chip buffers. The \textit{Top Control} module fetches and decodes instructions, before passing them to individual modules. As mentioned above in Section \ref{preprocessing}, the micro-architecture processes a single tile at a time.

\subsection{Overall Workflow}
\begin{figure}[tb]
    \centering
    \includegraphics[width=0.8\textwidth]{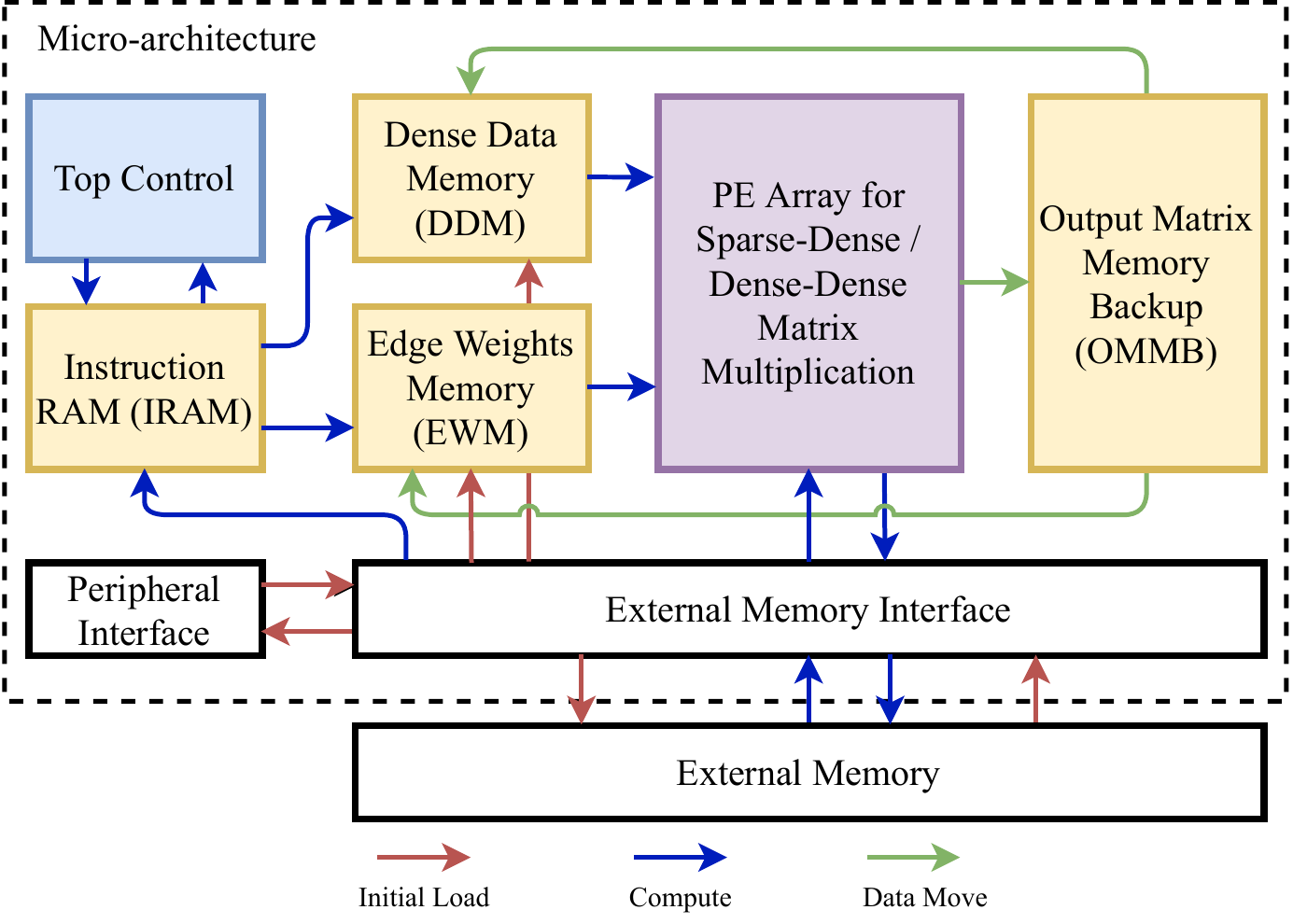}
    \caption{The overall micro-architecture and workflow of {\bf LW-GCN}}
    \label{fig:over_arch}
\end{figure}

The overall workflow of {\bf LW-GCN} is shown in Fig.~\ref{fig:over_arch}. The dense input data is transferred from external memory to \textit{dense data memory (DDM)} during the {\it initial load} step. During the {\it Compute} step, sparse input data is streamed onto the \textit{edge weights memory (EWM)}, and the {\it PE array} fetches $X$ data from \textit{EWM} and $W$ data from \textit{DDM} and performs multiply-accumulate (MAC) operations in parallel. Upon finishing all pairs of tiles from each {\it aggregation} or {\it combination} step, we move output to the \textit{output matrix memory backup (OMMB)}, and move a copy to {\it DDM} after {\it combination} and {\it EWM} after {\it aggregation} during the {\it data move} step.

\subsection{Multi-Bank Dense Data Memory (DDM)}
\label{subsec:ddm_par}
As mentioned in Section~\ref{preprocessing}, in order to compute different rows on different PEs in parallel, multiple non-zero elements from the sparse input are streamed on-chip during SDMM. Due to the sparseness and irregularity of $X$, it is difficult to predict the column positions of the non-zero elements ahead of time. Particularly, it is possible that several PEs require different addresses from the same {\it DDM}. Limited by read capability of on-chip memory (dual-port RAM only supports reading from two ports at most but the PE number is likely larger than two), such access restriction leads to data collision. In the micro-architecture of {\bf LW-GCN}, we build a multi-port memory through {\it data replication} and {\it row grouping} to reduce such data collision. In addition, we further reduce the occurrence of such data collision during preprocessing, as mentioned in Section~\ref{preprocessing}.

\subsubsection{Data replication} We replicate the dense data into $r$ replicas for different memory slices. Ideally, when setting $r$ equals to the number of PEs, the aforementioned data collision can be avoided because we would have a dedicated replica of dense data for each PE. However, this incurs a large on-chip memory requirement and is unfeasible in reality. Therefore, we set a relative small $r$ to solve part of the data collision with acceptable resource utilization (the chosen of $r$ is explained in detail in Section~\ref{hyper-parameter-impact}), and we introduce {\it row grouping} to further reduce the occurrence of data collision.

\subsubsection{Row grouping} We partition each dense data replica into $g$ row groups, each of which is stored independently. Specifically, we store row $W_j$ on group ($j$ mod $g$), so that data collision can only occur between elements $X_{i_a, j_a}$ and $X_{i_b, j_b}$ if $(j_a$ mod $g)$ = $(j_b$ mod $g)$ and $j_a \ne j_b$, which is significantly less likely compared with the undivided memory. Despite the fact that on-chip memory requires a minimum depth to be fully utilized, we are able to use high numbers of row groups to statistically reduce the probability of data collision. However, {\it row grouping} with large $g$ leads to high complexity for data distribution to PEs, which results in complex placement and routing and increases resource consumption. 

Both {\it data replication} and {\it row grouping} can efficiently reduce data collision. The remaining collision is avoided by injecting empty elements and processing them as idle cycles, as mentioned in section \ref{preprocessing}. We experiment with different $r$ and $g$ in section \ref{latency-breakdown}, where we analyze the number of inserted idle cycles versus hardware resource consumption to determine the optimal number of memory replicas and row groups. 

\begin{figure*}[tb]
    \centering
    \includegraphics[width=\textwidth]{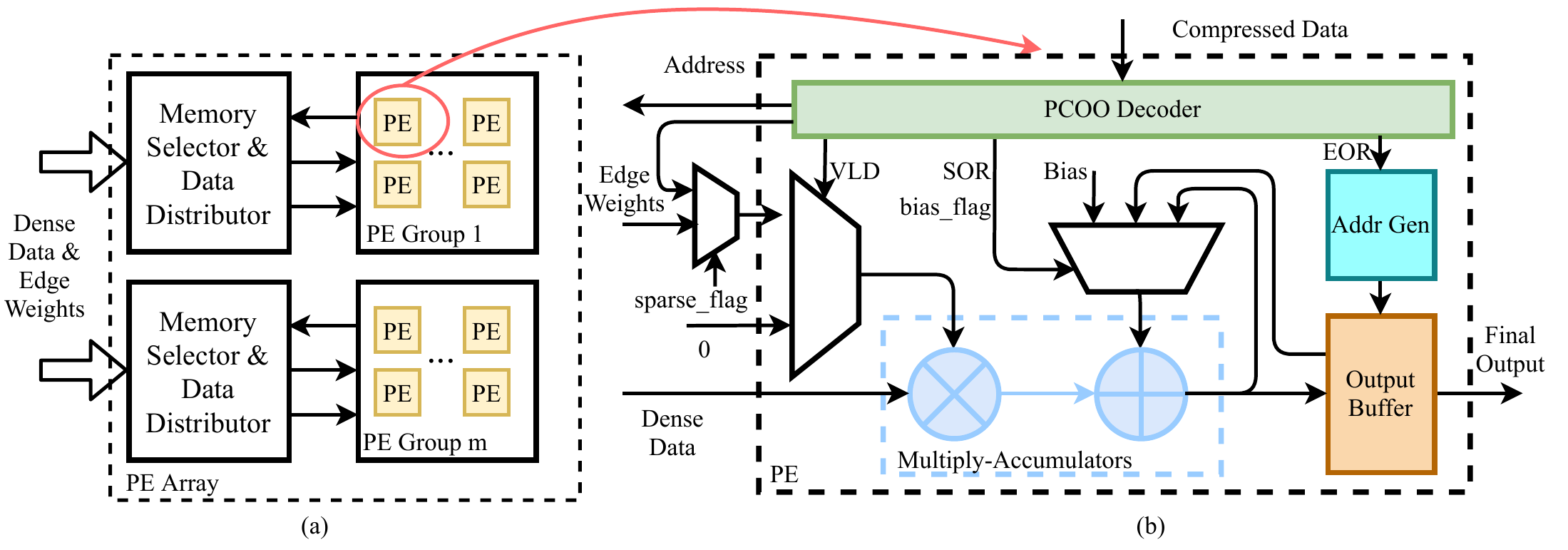}
    \caption{(a) The architecture of PE Array; (b) Detailed architecture of a PE.}
    \label{fig:pe_array}
\end{figure*}

\subsection{Unified PE architecture for DMM and SDMM}
As shown in Fig.~\ref{fig:pe_array} (a), the number of PE groups and memory banks are kept the same, so that each PE group can access the corresponding memory bank for dense data to avoid data collision. Based on addresses generated by an individual PE, {\it Memory Selector} and {\it Data Distributor} dispatch appropriate dense data. We use priority decoder when distributing addresses to memory banks, which allows different PEs to fetch from the same address of the same memory bank.

Note that data replication only applies to dense input but not sparse input. The compressed sparse data is streamed directly to each PE. As shown in Fig.~\ref{fig:pe_array}(b), data first passes through {\it PCOO Decoder}, where the $log_2(T)$-bit column index is interpreted as memory address to fetch dense data. If a valid bit is observed (VLD = 1), the PE routes the corresponding value to its multiplier, otherwise it assumes the current value is an injected empty element (i.e. data collision, waiting for other PEs to finish, etc.), and routes 0 to the multiplier instead. Since multiple rows are concatenated to feed into each PE, we use SOR and EOR to indicate the start and end of a row, respectively. For each computation step, SOR controls the input of the accumulator to be either its previous result (SOR = 0) or the intermediate result of the previous tile saved in OMMB (SOR = 1). Meanwhile, EOR controls the address generation for storing current results into the output buffer, and also increments the internally tracked row number (EOR = 1).

The DMM is also performed in the PE with the same working flow. Since the left matrix is dense, all the rows share the same row and column information, which also goes through the PCOO decoder. The sparse\_flag signal then indicate which data to select. When we are processing DMM (sparse\_flag is 0), we will select the edge weights stored in EWM, otherwise, we will select the value decoded from {\it PCOO Decoder}. In this way, we can perform both DMM and SDMM in the unified PE, which increases the working efficiency of PE for computing {\it combination} and {\it aggregation} of GCNs.

\begin{figure*}[tb]
    \centering
    \includegraphics[width=0.45\textwidth]{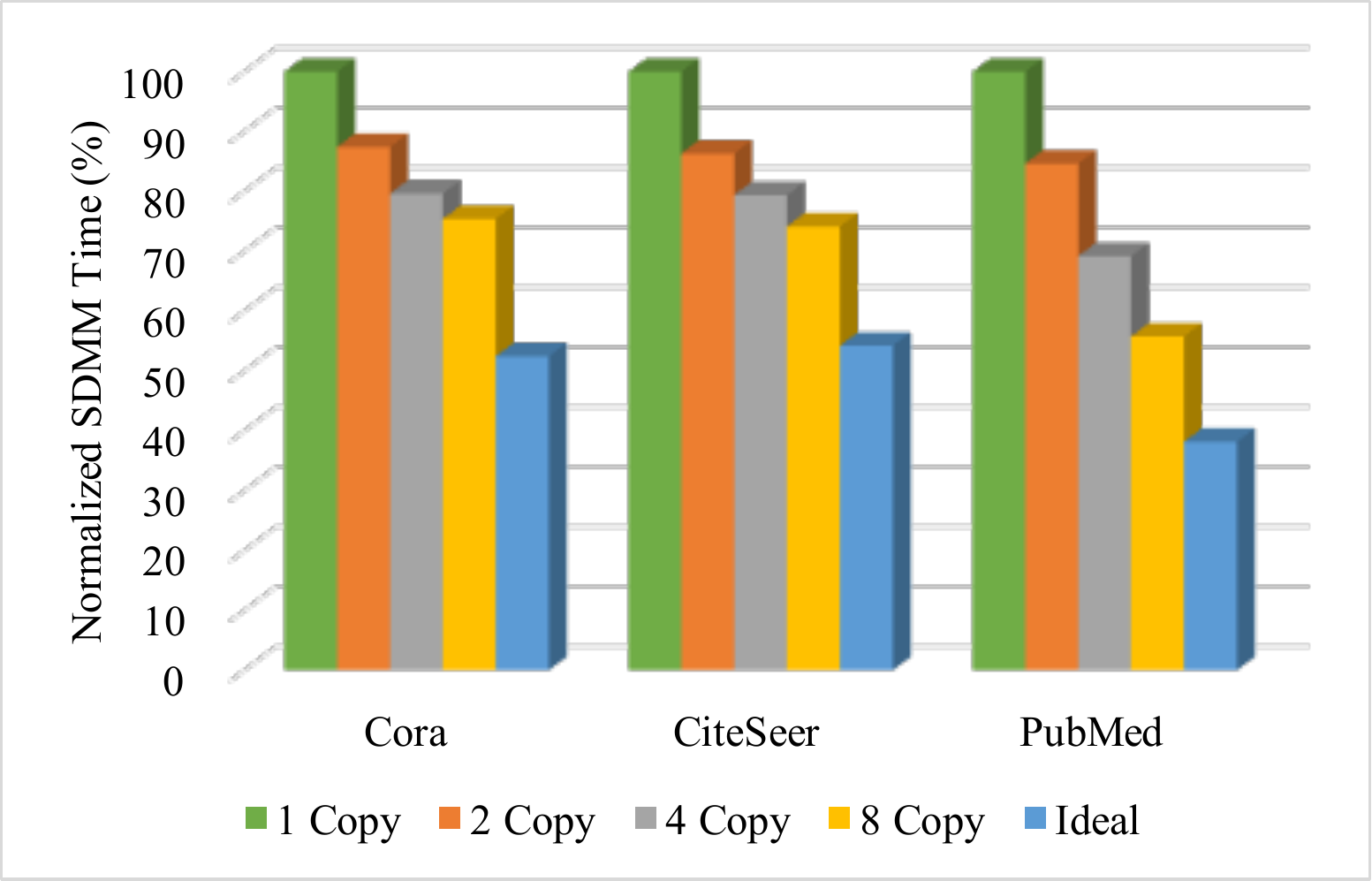}
    \includegraphics[width=0.45\textwidth]{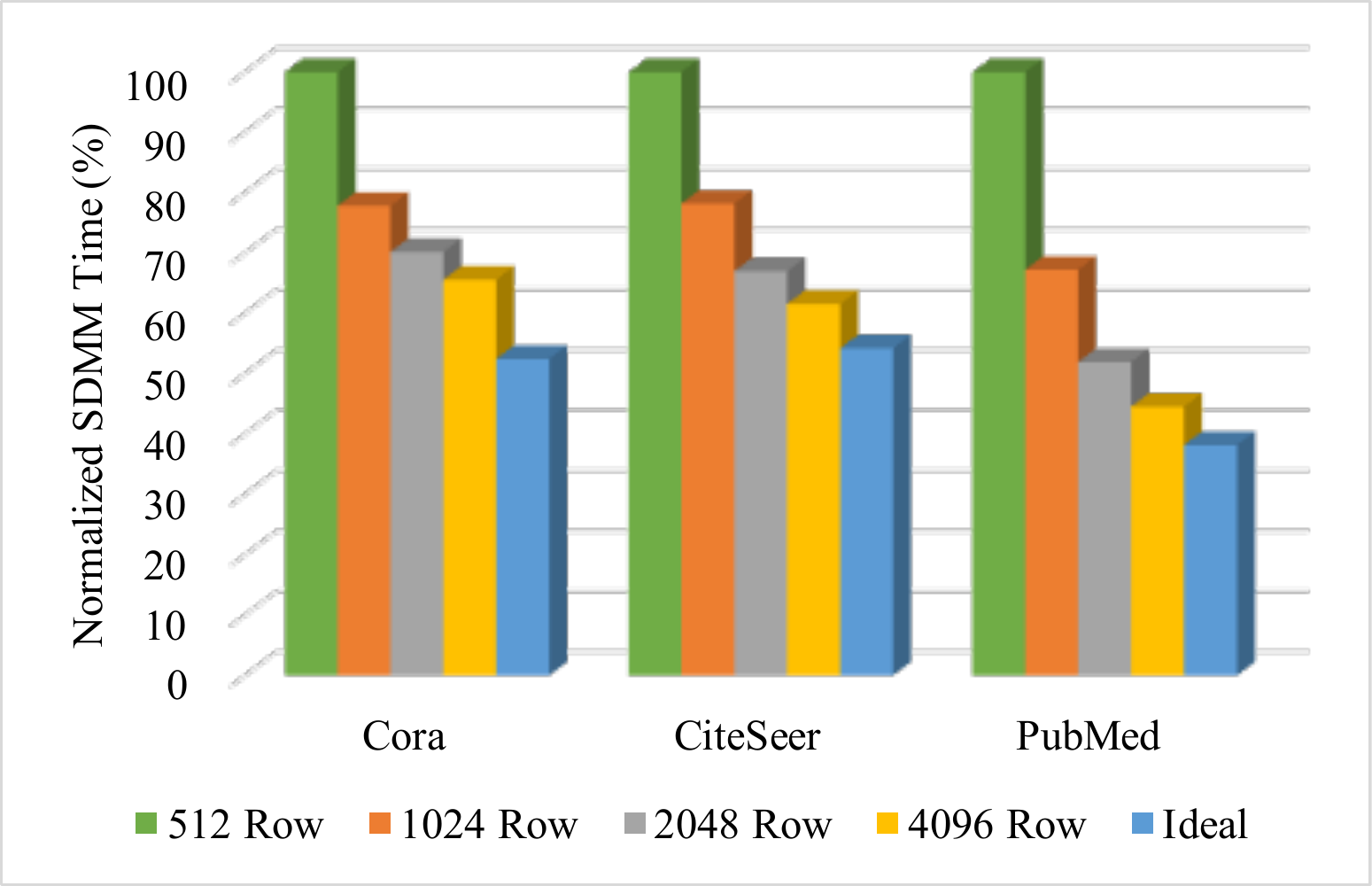}
    \caption{Impact of (a) dense data replication with 512-row tiles and (b) tile size with 1 replica}
    \label{replication-impact}
\end{figure*}

\section{Evaluation}
\label{evaluation}

In this section, we evaluate {\bf LW-GCN} on different configurations to identify the impact of each hardware resource. We then compare a final implementation against existing computing platforms on three popular datasets: Cora, CiteSeer, and PubMed. The dimensions and densities of each dataset are shown in table \ref{dataset-details}.

\subsection{Experiment Design}
We evaluate \textbf{LW-GCN} on a two-layer GCN which use a hidden size of 16 and trained dense weights and bias via the state-of-art framework Pytorch Geometric (PyG). Note that this setup is identical to the GCN used in \cite{cite-awb-gcn} which we will be evaluating against. In addition, to demonstrate the flexibility of our approach, we extend our evaluation to GraphSAGE under the same datasets.

{\bf LW-GCN} is implemented in Verilog HDL and deployed onto a Xilinx Kintex-7 K325T FPGA, where we measure the execution time and energy consumption. The DDM is implemented with LUT RAM while other on-chip memories are implemented with Block RAM (BRAM). This is because memory banks in DDM require small depth and high bandwidth, and LUT RAM is more suitable than BRAM. In this section, we first explore the impact of tile size and dense input replication on execution latency. Then, we present a breakdown of latency in individual step of loading, computation, and data movement. Finally, we present an overall performance comparison against existing platforms in terms of latency and energy efficiency.

\subsection{Hyper Parameter Impact}
\label{hyper-parameter-impact}
During each SDMM step, the dense input is stored in on-chip LUT RAM, where multiple rows would be stored on the same slice of memory in order to fully utilize it. The limitation where only a single row can be read from each LUT RAM slice at a time induces data collision when multiple reads are needed for a same RAM slice and at a same time. As explained in Section~\ref{subsec:ddm_par}, both {\it data replication} and {\it row grouping} can effectively reduce data collision. The less data collision will in return results in smaller latency of computing SDMM. On the other hand, due to the irregular nature of graph adjacency matrices, individual rows have very different sparsity which results in PE imbalance, we statistically minimize this effect by utilizing larger tiles. As GCN has the hidden size of 16, we set each PE to have 16 multiply-accumulators and have the fixed relationship between tile size $T$ and row grouping $g$ that $T=16g$. Therefore, we evaluate the impact of latency from dense data replication $r$ and tile size $T$, as shown in Fig.~\ref{replication-impact}. We can see that the latency of computing is decreased by more dense data replications as well as larger tile sizes. At 8 replicas, \textbf{LW-GCN}'s SDMM latency is reduced by up to 44.23\% (on PubMed) compared to 1 replica under the same 512-row tile setup. At 4096-row tiles, SDMM latency is reduced by up to 61.83\% (on PubMed) with the same replication setup. The ideal cases in Fig.~\ref{replication-impact} is estimated by summing up the total amount of workload, and assuming every PE is fully utilized.

\begin{figure*}[tb]
    \centering
    \includegraphics[width=0.47\textwidth]{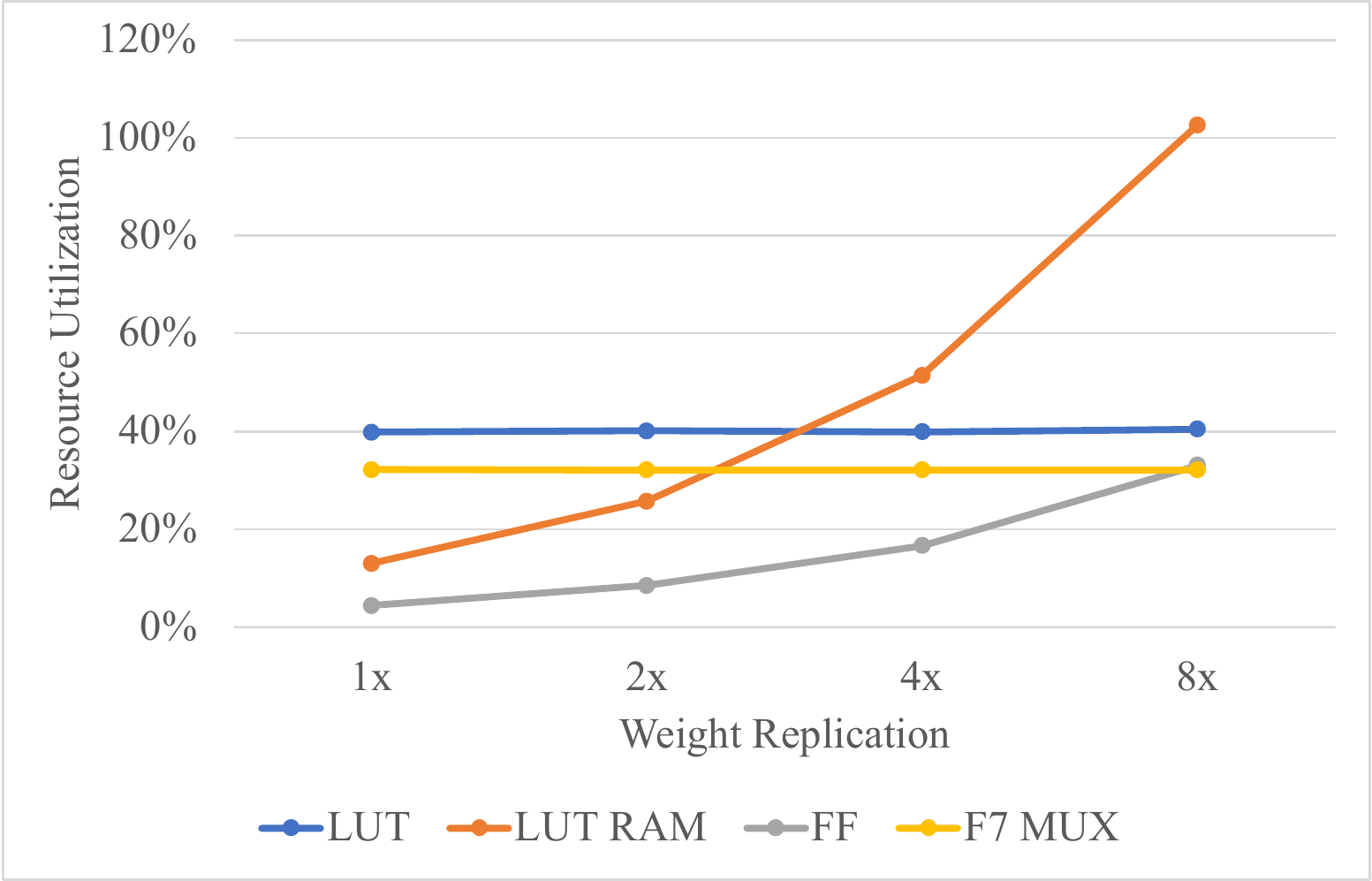}
    \includegraphics[width=0.47\textwidth]{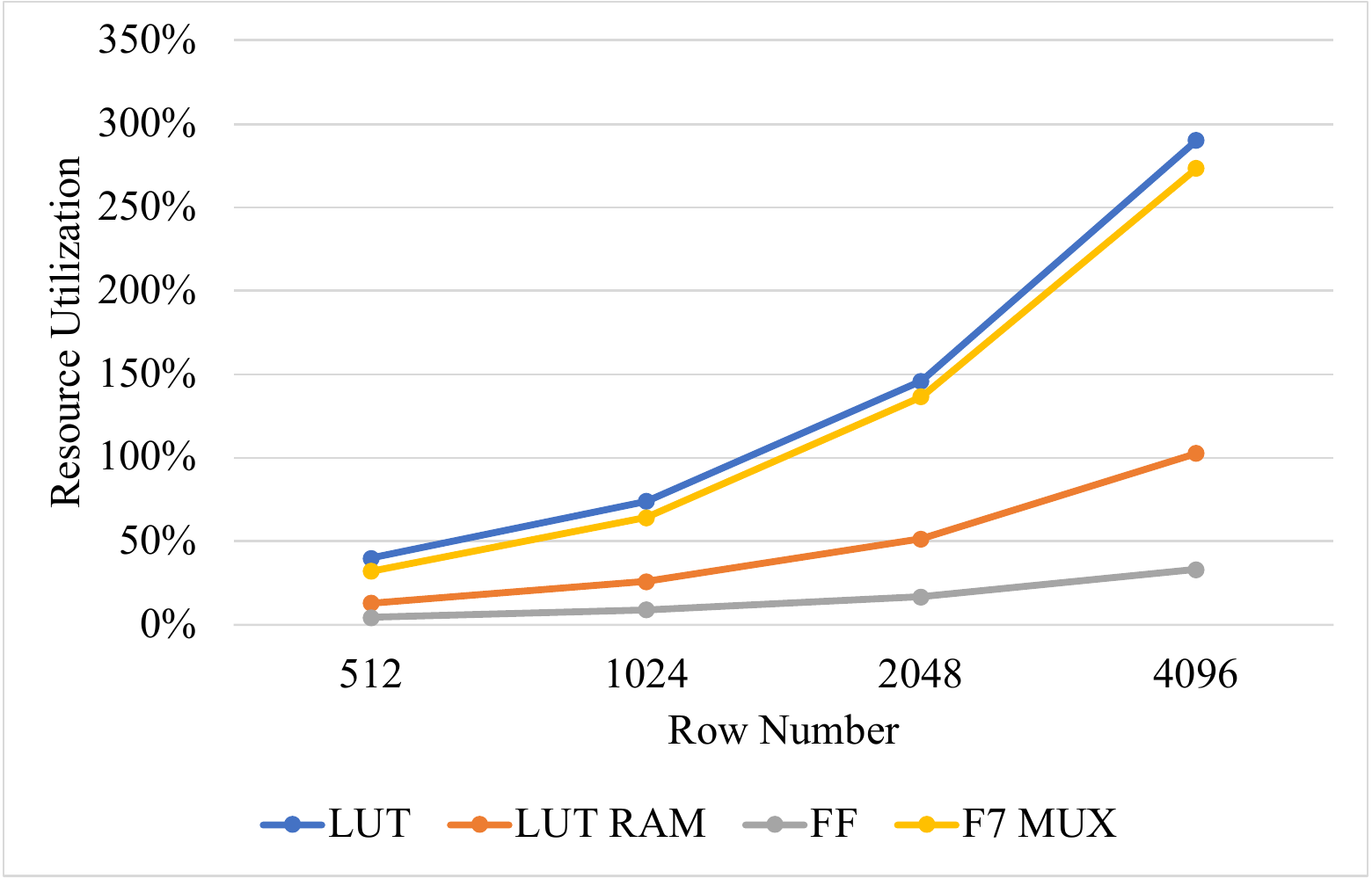}
    \caption{Resource consumption of (a) replication and (b) tile size}
    \label{fig:replica-consumption}
\end{figure*}

Due to resource limitations, it is unfeasible to continuously expand tile sizes and replication numbers. In this way, we evaluate the resource utilization under the resource limited device with respect to dense data replication $r$ and tile size $T$, as shown in Fig.~\ref{fig:replica-consumption}. According to the results in Fig.~\ref{fig:replica-consumption}, $r=4$ and $T=512$ ($g=32$) achieve the best balance resource and performance under the specific FPGA, and will be used for the remaining of experiments. Given these hyper parameters, the overall resource utilization on Kintex-7 K325T FPGA is shown in Table \ref{fpga-resource}.

\begin{table*}[tb]
    \centering
    \caption{Resource utilization on Kintex-7 325T FPGA.}
    \label{fpga-resource}
    \begin{tabular}{l|c|c|c|c|c|c} \hline\hline
        Resource            & LUT       & LUT RAM  & FF     & F7 MUX    & BRAM      & DSP       \\ \hline
        Used                & 161529    & 33804    & 94369  & 32768     & 291.5     & 512       \\ \hline
        Available           & 203800    & 64000    & 407600 & 101900    & 445       & 840       \\ \hline
        Utilization (\%)    & 79.26     & 52.82    & 23.15  & 32.16     & 65.51     & 60.95     \\ \hline \hline
    \end{tabular}
\end{table*}

\subsection{Latency Breakdown}
\label{latency-breakdown}
\begin{figure*}[tb]
    \centering
    \includegraphics[width=0.45\textwidth]{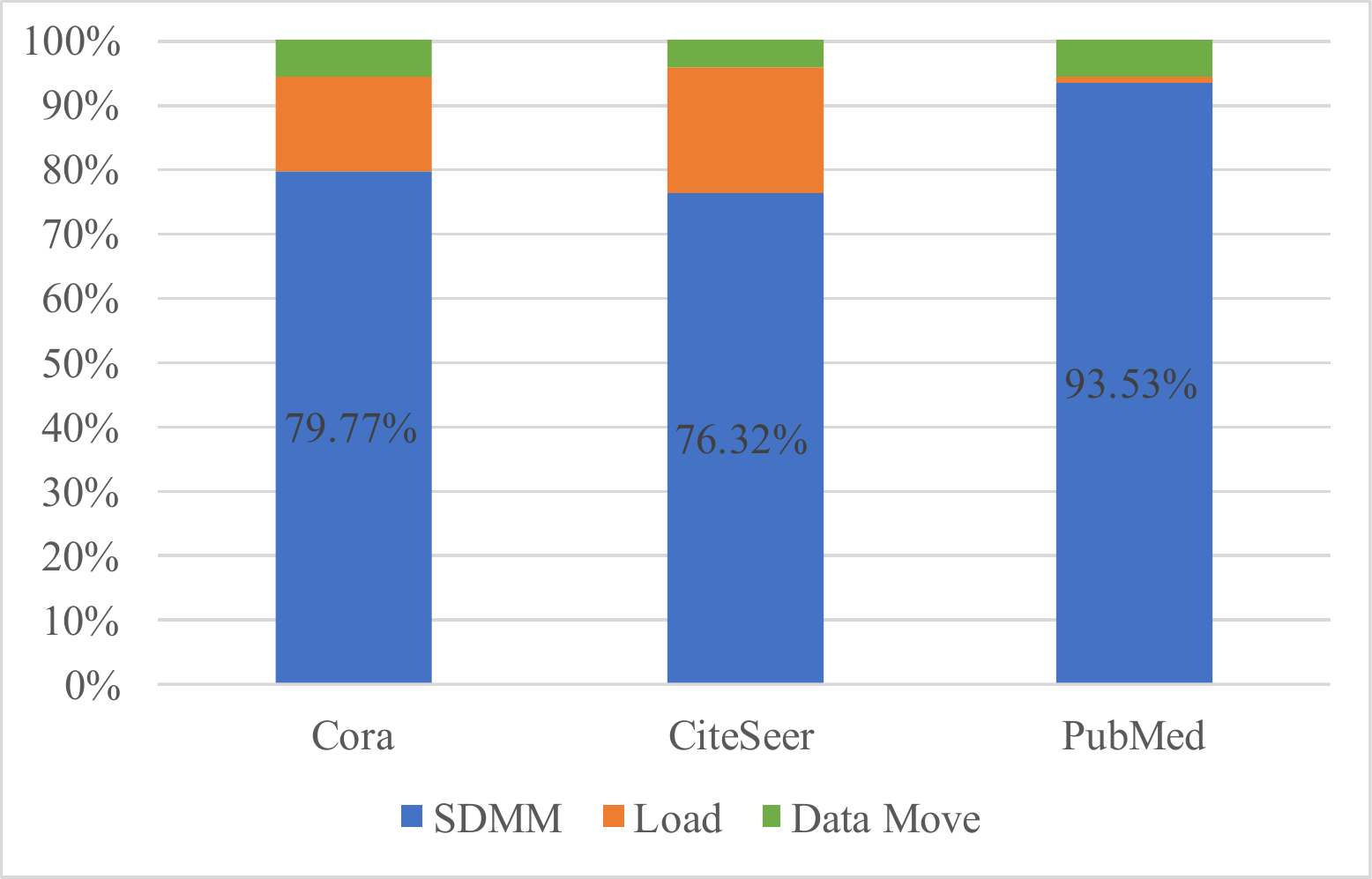}
    \includegraphics[width=0.45\textwidth]{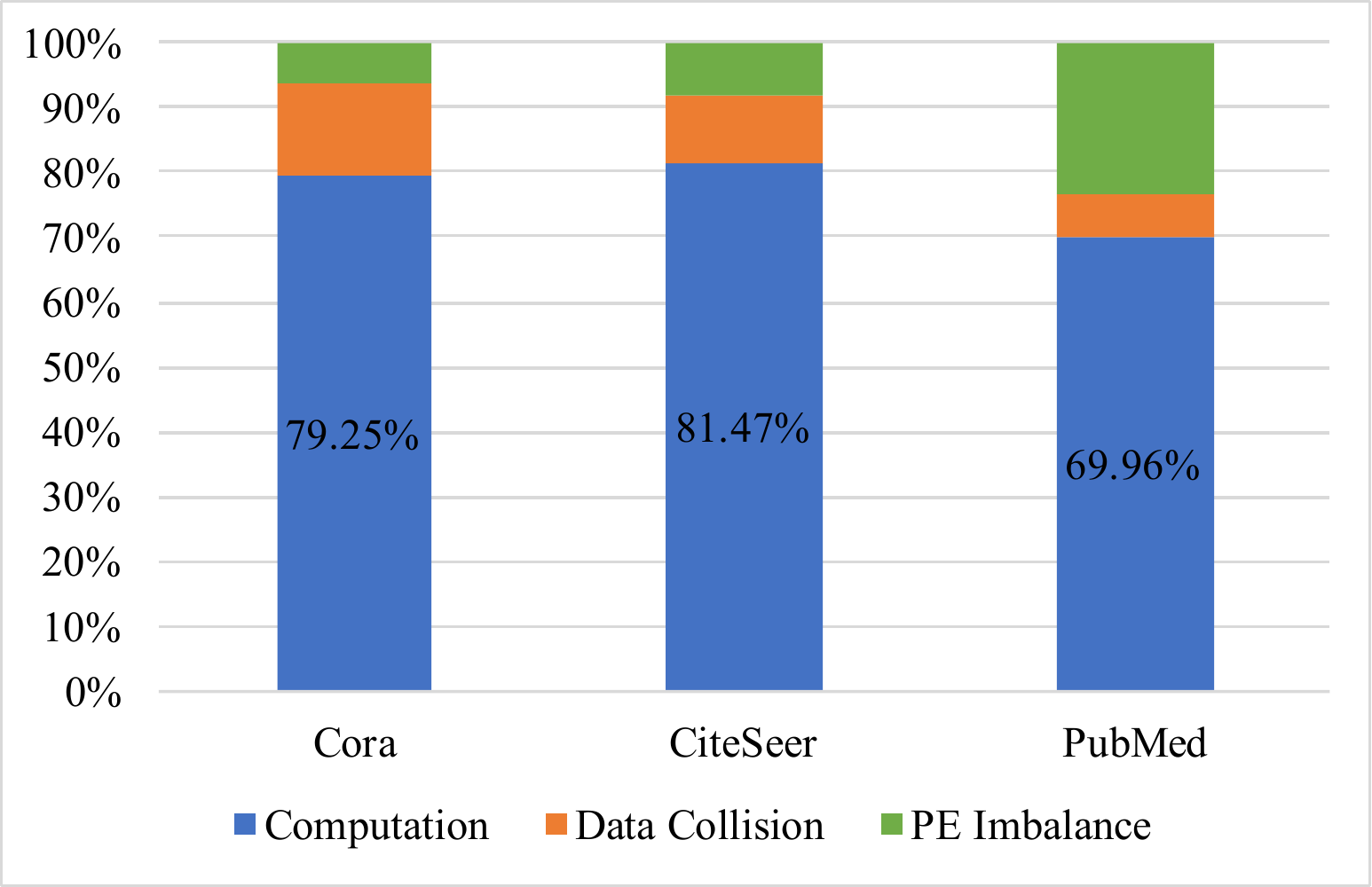}
    \caption{Latency breakdown for (a) full execution and (b) SDMM}
    \label{latency}
\end{figure*}

During preprocessing, we inject empty elements (see section~\ref{assignment_scheduling}) to handle the corner case where a row $X_i$ contains no non-zero elements, or when a PE completes its execution. This enables each PE to internally track current row $i$, which allows us to remove row number $i$ from off-chip memory and reduce memory bandwidth consumption. We also inject empty element symbols when two elements are to read from different depths of the same memory, in order to prevent data collision. Fig.~\ref{latency} shows the latency breakdown for overall runtime (including DMM/SDMM, memory load and on-chip data movement) as well as for SDMM (including computation, PE imbalance, and data collision). The latency of DMM is dominant by computation as DMM does not have the issues of PE imbalance and data collision. Therefore, we do not list the latency breakdown for DMM. In both cases, the time spending on computation ({\it i.e.} DMM/SDMM for overall, and computation for SDMM) is dominant. The dataset PubMed has a relatively larger PE imbalance. This indicates potential room for improvement by workload scheduling, which will be explored in the future. 

\begin{figure*}[tb]
    \centering
    \includegraphics[width=0.45\textwidth]{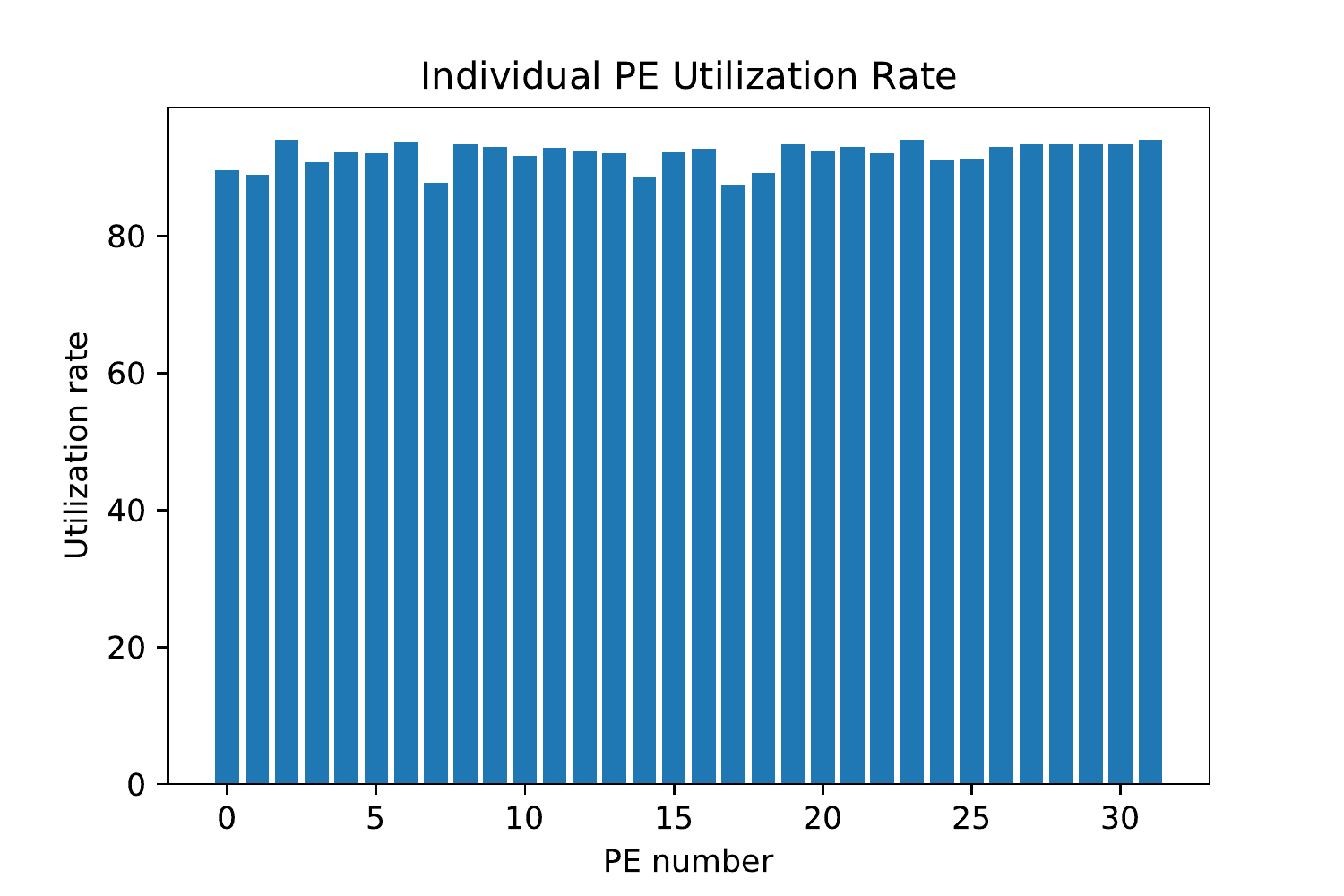}
    \includegraphics[width=0.45\textwidth]{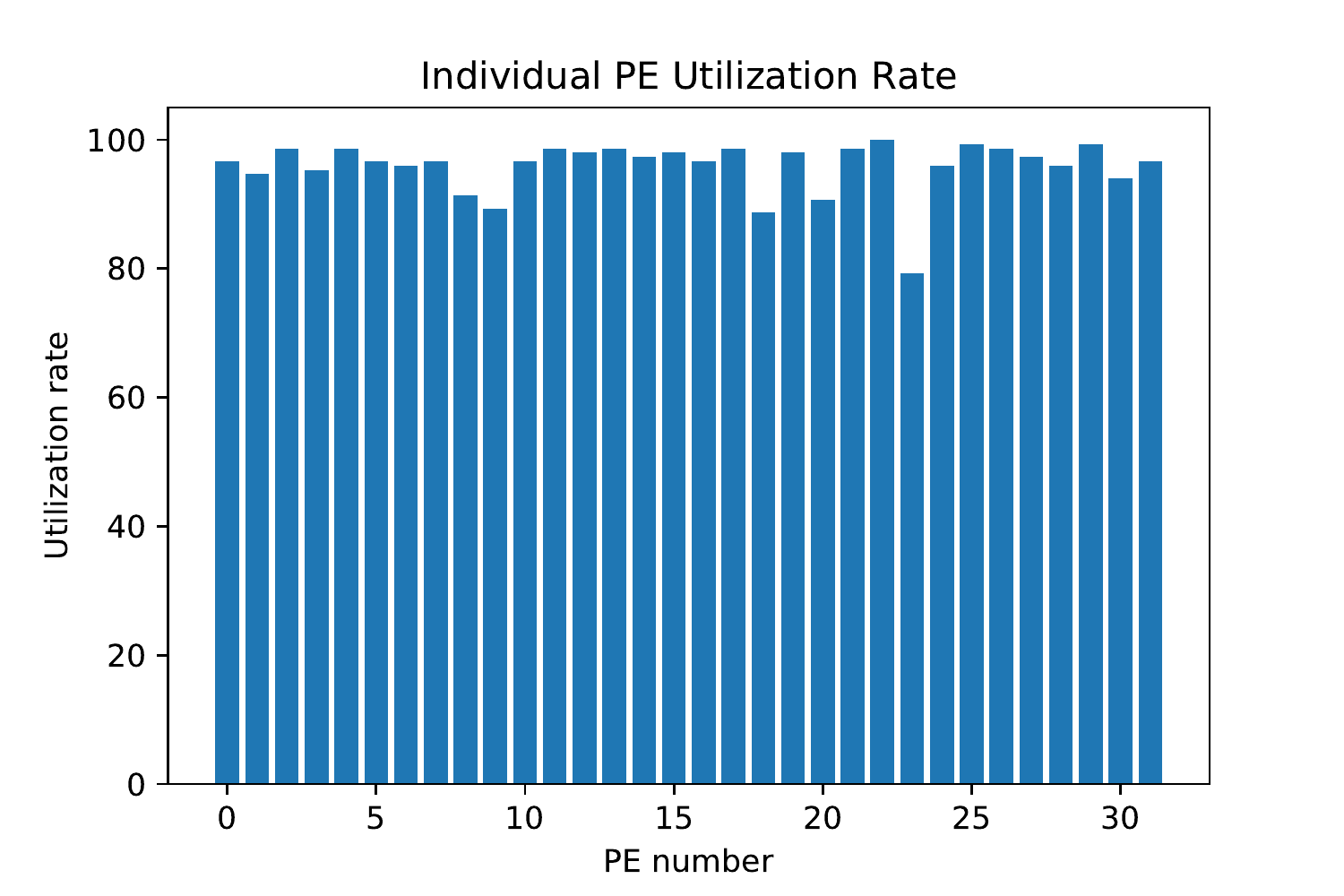}
    \caption{PE utilization during SDMM for Cora: (a) first combination tile and (b) first aggregation tile}
    \label{pe-utilization}
\end{figure*}

We further evaluate the specific utilization rates per PE with respect to {\it combination} and {\it aggregation} operations. For simplicity we only show the first layer on Cora dataset in Fig.~\ref{pe-utilization}. It shows that the idle time of each PE varies from 6\% to 12\% for {\it combination} and from 1\% to 20\% for {\it aggregation}, respectively. In overall, the lowest utilized PE is idle for less than 20\% of the SDMM time. 

\subsection{Overall Comparison}
We evaluate the overall latency and energy efficiency of {\bf LW-GCN} against the Intel Xeon Gold 5218 CPU, NVIDIA Xavier NX edge GPU, NVIDIA RTX3090 GPU, and state-of-the-art FPGA-based GCN accelerator AWB-GCN~\cite{cite-awb-gcn}, and report results in the top half (GCN) of Table~\ref{overall-comparison}. Note that AWB-GCN is implemented on Intel Stratix 10 D5005 with frequency of 330 MHz and uses 8192 DSP slices. We normalize their reported latency and energy efficiency with this resource utilization to our FPGA (200 MHz and 512 DSP slices) for a fair comparison. For energy efficiency, Intel Stratix 10 D5005 uses 14nm transistors while Xilinx Kintex-7 325T uses 28nm transistors, following the analysis in \cite{cite-energy}, we normalize their power consumption by $(\frac{28}{14})^2=4\times$. For GCN as illustrated in Table~\ref{overall-comparison}, {\bf LW-GCN} outperforms all the other platforms in terms of latency and energy efficiency. Specifically, {\bf LW-GCN} achieves up to 60$\times$, 32$\times$, 12$\times$ and 1.7$\times$ speedup, as well as 2478$\times$, 84$\times$, 511$\times$, and 3.88$\times$ energy efficiency, compared with CPU, edge GPU, GPU, and AWB-GCN, respectively. \textbf{LW-GCN} is able to achieve such performance benchmarks while keeping a small resource budget, due to the techniques used in software preprocessing and micro-architecture to reduce data collision and PE imbalance for SDMM, as well as performing DMM and SDMM with unified architecture.

\begin{table*}[tb]
    \centering
    \footnotesize
    \caption{Comparison with CPU, edge GPU, general GPU and existing FPGA accelerator on GCN and GraphSAGE}
    \label{overall-comparison}
    
    \begin{tabular}{l|c|c|c|c|c|c} \hline\hline
        & \multicolumn{3}{c|}{Latency (ms) [speedup]} & \multicolumn{3}{c}{Energy efficiency (graph/kJ)} \\
        \cline{2-7}
        Platform & Cora & CiteSeer & PubMed & Cora & CiteSeer & PubMed \\ \cline{2-7}
        (Clock rate: GHz)& \multicolumn{6}{c}{GCN} \\ \hline
        Intel Xeon Gold 5218 (2.1) & 1.89 [1$\times$] & 3.88 [1$\times$] & 12.5 [1$\times$] & 4.23E3 & 2.06E3 & 640 \\ \hline
        NVIDIA Xavier NX (1.1) & 1.87 [1$\times$] & 1.88 [2.1$\times$] & 2.01 [6.2$\times$] & 3.57E4 & 3.55E4 & 3.32E4 \\ \hline
        NVIDIA RTX3090 (1.7) & 0.492 [3.9$\times$] & 0.481 [8.1$\times$] & 0.491 [26$\times$] & 5.83E3 & 5.95E3 & 5.83E3 \\ \hline
        AWB-GCN (0.2) & 0.0613 [31$\times$] & 0.115 [35$\times$] & 0.791 [16$\times$] & 7.70E5 & 4.82E5 & 6.21E5 \\ \hline
        {\bf LW-GCN (0.2)} & {\bf 0.0412 [46$\times$]} & {\bf 0.0652 [60$\times$]} & {\bf 0.571 [22$\times$]} & {\bf 2.98E6} & {\bf 1.88E6} & {\bf 2.14E5} \\ \hline \hline
        & \multicolumn{6}{c}{GraphSAGE} \\ \hline
        Intel Xeon Gold 5218 (2.1) & 172 [1$\times$] & 385 [1$\times$] & 340 [1$\times$] & 46.5 & 20.8 & 23.5 \\ \hline
        NVIDIA Xavier NX (1.1) & 10.6 [16.3$\times$] & 9.63 [40.0$\times$] & 10.8 [31.5$\times$] & 6.28E3 & 6.92E3 & 6.17E3 \\ \hline
        NVIDIA RTX3090 (1.7) & 1.94 [89.0$\times$] & 1.88 [204.8$\times$] & 1.96 [173.6$\times$] & 1.47E3 & 1.52E3 & 1.46E3 \\ \hline
        AWB-GCN (0.2) & NA & NA & NA & NA & NA & NA \\ \hline
        {\bf LW-GCN (0.2)} & {\bf 0.086 [2.01E3$\times$]} & {\bf 0.14 [2.75E3$\times$]} & {\bf 1.07 [318$\times$]} & {\bf 1.42E6} & {\bf 8.77E5} & {\bf 1.72E4} \\ \hline \hline
    \end{tabular}
\end{table*}

\subsection{Extending LW-GCN to Other Algorithms}
Although \textbf{LW-GCN} is designed as a GCN accelerator, the underlying DMM/SDMM acceleration is not limited to GCN, and can be applied to any DMM/SDMM related GNN workloads. In fact, due to the sparse nature of graph adjacent matrices and dense nature of weight matrices, most GNNs workload involves DMM/SDMM. As a proof of concept we directly applied {\bf LW-GCN} to GraphSAGE~\cite{cite-graphsage} on the same datasets, and achieved an acceleration of up to 2750x, 123x, and 22.6x and energy savings of up to 42200x, 226x, and 966x over CPU, edge GPU, and GPU respectively, as shown in the bottom half of Table~\ref{overall-comparison}. Note that AWB-GCN results for GraphSAGE is not available in the literature. Additionally, note that the PyG implementation for GraphSAGE involves a sparse-sparse matrix multiplication due to computing \textit{aggregation} before \textit{concatenation}, when applied on the three datasets we used, therefore the latency is much higher than it could be.

\section{Conclusions and Future Work}
GCN involves heavy computation of multiplications of sparse and dense matrices, but most neural network accelerators are targeted at CNN with dense matrix multiplication and therefore are not efficient for GCN. Recently, FPGA-based {\bf AWB-GCN} improves performance, but still requires a large amount of on-chip memory. Therefore, it is inapplicable to resource limited hardware platforms such as edge devices.

In this paper, we have proposed {\bf LW-GCN}, a software-hardware co-designed accelerator for GCN inference. {\bf LW-GCN} consists of a software preprocessing algorithm and an FPGA-based hardware accelerator. The core to {\bf LW-GCN} is our SDMM design, which reduces memory needs through tiling, data quantization, sparse matrix compression, and workload assignment with data collision resolution. Experiments show that for GCN, {\bf LW-GCN} reduces latency by up to 60x, 12x, and 1.7x compared to CPU, GPU, and AWB-GCN and increases power efficiency by up to 912x, 511x, and 3.87x. Additionally, the underlying SDMM design used by {\bf LW-GCN} is applicable to other graph neural network algorithms such as GraphSAGE, not limited to GCN. 


\bibliographystyle{ACM-Reference-Format}
\bibliography{main}


\end{document}